\documentclass[conference]{IEEEtran}
\IEEEoverridecommandlockouts
\usepackage{amsmath}
\usepackage{cite}
\usepackage{hyperref}
\usepackage{graphicx}
\usepackage{booktabs}
\usepackage{tikz}
\usetikzlibrary{calc}
\usepackage{dcolumn}
\usepackage{xcolor}
\newcolumntype{d}[1]{D{.}{.}{#1}}

\begin{document}

\title{Perceiver\textit{S}: A Multi-\textit{S}cale Perceiver with Effective \textit{S}egmentation for Long-Term Expressive Symbolic Music Generation}

\author{
    \IEEEauthorblockN{Yungang Yi\IEEEauthorrefmark{1}, Weihua Li\IEEEauthorrefmark{1}, Matthew Kuo\IEEEauthorrefmark{1}, Quan Bai\IEEEauthorrefmark{2}}
    \IEEEauthorblockA{\IEEEauthorrefmark{1}Auckland University of Technology, Auckland, New Zealand\\
    Email: yungang.yi@aut.ac.nz, weihua.li@aut.ac.nz, matthew.kuo@aut.ac.nz}
    \IEEEauthorblockA{\IEEEauthorrefmark{2}University of Tasmania, Tasmania, Australia\\
    Email: quan.bai@utas.edu.au}

    \thanks{© 2025 IEEE.  Personal use of this material is permitted. 
    Permission from IEEE must be obtained for all other uses, in any current or future media, including reprinting/republishing this material for advertising or promotional purposes, creating new collective works, for resale or redistribution to servers or lists, or reuse of any copyrighted component of this work in other works.  
    DOI: \href{https://doi.org/10.1109/TASLPRO.2025.3611836}{10.1109/TASLPRO.2025.3611836}}
}

\maketitle

\begin{abstract}
AI-based music generation has made significant progress in recent years. However, generating symbolic music that is both long-structured and expressive remains a significant challenge. In this paper, we propose Perceiver\textit{S} (\textit{S}egmentation and \textit{S}cale), a novel architecture designed to address this issue by leveraging both \textbf{Effective \textit{S}egmentation} and \textbf{Multi-\textit{S}cale} attention mechanisms. Our approach enhances symbolic music generation by simultaneously learning long-term structural dependencies and short-term expressive details. By combining cross-attention and self-attention in a \textbf{Multi-\textit{S}cale} setting, Perceiver\textit{S} captures long-range musical structure while preserving performance nuances. The proposed model has been evaluated using the Maestro dataset and has demonstrated improvements in generating coherent and diverse music, characterized by both structural consistency and expressive variation. The project demos and the generated music samples can be accessed through the link: \url{https://perceivers.github.io}. 
\end{abstract}


\section{Introduction}

Recent advancements in AI-based music generation, especially in audio generation models such as AudioLDM~\cite{liu2023audioldmtexttoaudiogenerationlatent}, MusicGen~\cite{copet2024simplecontrollablemusicgeneration}, and Jen-1~\cite{li2023jen1textguideduniversalmusic}, have demonstrated significant progress capable of generating highly natural-sounding music. Compared to audio generation, symbolic music offers a higher level of abstraction, making it easier for machine learning models to capture and understand deeper musical characteristics. Symbolic music plays a crucial role in the field of music generation, particularly due to its editable nature. This allows for operations such as cutting and rearranging different sections or substituting instrument timbres, enabling human involvement in high-quality music production during the post-processing stage. 

However, symbolic music generation faces two key challenges. \textbf{First}, the most expressive datasets, recorded from live performances and recording studios, are seldom used compared to manually created MIDI-file datasets. The primary reason is that they lack detailed annotations, making it harder for models to learn their complex structures. Furthermore, due to computational limitations, these models cannot fully capture the context of an entire piece of music. Techniques, such as \textit{chunking} and \textit{quantization}, are often employed to reduce computational complexity, leading to the loss of crucial musical details and making it difficult for models to grasp the full structure of a composition. \textbf{Second}, other approaches that use abstract structural representations as conditions for music generation tasks have enabled the generation of structured music. However, such methods rely heavily on handcrafted feature engineering. Our objective is to design and develop a model that is capable of learning the long-range dependencies in music without relying on explicit structural annotations.

The emergence of Transformer Attention technologies, such as \textit{Perceiver AR}~\cite{hawthorne2022generalpurposelongcontextautoregressivemodeling}, has made it possible to access much longer contextual dependencies. It allows for the simultaneous learning of musical structure and the generation of expressive performances. Perceiver AR has demonstrated the ability to attend to a context length of up to \textit{32,768 tokens} using the Maestro dataset~\cite{hawthorne2019enablingfactorizedpianomusic}, where the query in cross-attention attends to significantly longer key/value pairs ~\cite{hawthorne2022generalpurposelongcontextautoregressivemodeling}.

In the Transformer attention architecture, a \textit{causal mask} is a key mechanism that prevents each token from accessing future tokens during training. This prevents information leakage and enforces autoregressive behavior. A detailed explanation of this mechanism is provided in Section~\ref{sec:preliminaries}.

However, the \textit{causal mask}, when applied with the default input sequence segmentation, does not fully conceal tokens that should not be visible during auto-regressive training and generation, which ultimately degrades the quality of the generated music. Additionally, when using ultra-long context as a condition, the model tends to generate identical or similar repetitive segments as the sequence length increases due to issues with \textit{high similarity} in the context of neighboring tokens, leading to a high token autocorrelation~\cite{lee2023mathematical} tendency.

In this paper, we propose Perceiver\textit{S}, a novel model that addresses the causal masking issue by incorporating Effective Segmentation. Additionally,  Perceiver\textit{S} employs Multi-\textit{S}cale attention to mitigate the high token autocorrelation problem that arises from relying solely on long-range dependencies. Specifically, by adjusting the input sequence segmentation to start from the head segment with an effective causal mask and aggressively increasing the segment length up to the maximum input sequence length, we resolve the learning limitations caused by the causal mask in Perceiver AR~\cite{hawthorne2022generalpurposelongcontextautoregressivemodeling}. Additionally, by incorporating Multi-Scale masks across multiple layers of cross-attention, the model considers both ultra-long and short-range attention simultaneously. This approach addresses the limitation in Perceiver AR, which focuses solely on long-range attention~\cite{hawthorne2022generalpurposelongcontextautoregressivemodeling}. Different from Perceiver AR, our approach enhances symbolic music generation by effectively capturing both long-term structural dependencies and short-term expressive details. Through improved segmentation and multi-scale attention mechanisms, Perceiver\textit{S} generates coherent, diverse music without relying on extensive structural annotations. Extensive experiments have been conducted to evaluate the performance of the proposed Perceiver\textit{S}. The experimental results demonstrate an average \textbf{40\%} improvement in Overlap Area when measured against the original training dataset, highlighting a substantial advantage of our approach over Perceiver AR~\cite{hawthorne2022generalpurposelongcontextautoregressivemodeling} in generating high-quality symbolic music.

\section{Related Works} \label{sec:related_works}

\subsection{Capturing Music Expressiveness}
The most significant early work in this area came from Google's team, which introduced \textbf{Music Transformer}~\cite{huang2018musictransformer}, a model capable of generating expressive piano music using the Piano-e-Competition dataset, later known as the \textbf{Maestro} dataset~\cite{hawthorne2019enablingfactorizedpianomusic}. Since this model was trained using MIDI files recorded from live piano performances, it utilized the attention mechanism to focus on the context of all previously generated tokens when predicting the next one, allowing it to generate highly detailed and expressive music. 

However, due to the \textit{quadratic complexity} \(O(n^2)\) of the Transformer attention mechanism, it could only generate music lasting for tens of seconds, but not for several minutes. Unlike Music Transformer~\cite{huang2018musictransformer}, few other models utilize performance datasets, primarily due to the lack of annotations associated with these types of datasets and the computational limitations involved in processing them.

\subsubsection{Dataset Selection}
A key factor in generating production-quality music lies in the selection of datasets. The \textbf{Maestro dataset}~\cite{hawthorne2019enablingfactorizedpianomusic}, consisting of real human performances, offers dynamic and expressive recording music. Datasets recorded from live performances are rare, but there are quite a few Automatic Music Transcription (ATM) datasets, including GiantMIDI~\cite{kong2022giantmidipianolargescalemididataset}, ATEPP~\cite{zhang2022atepp}, PiJAMA~\cite{edwards2023pijama}, and others. Advanced models have been developed, from Hawthorne~\cite{hawthorne2018onsetsframesdualobjectivepiano} to Kong~\cite{kong2022giantmidipianolargescalemididataset}, that convert audio into MIDI files. These advances have made it possible to use a vast amount of recorded audio music to train models, as the development of symbolic music models has long been constrained by the limited availability of datasets.

However, existing publicly available performance datasets are limited to solo piano recordings. There are currently no open datasets featuring solo performances of other instruments or multi-instrument ensemble performances.

Manually created MIDI datasets, like LAKH~\cite{raffel2016learning}, provide valuable human-annotated information, e.g., beats, bars, and phrases, which allows for flexible segmentation and richer feature extraction but lack expressive nuances found in live performances, such as dynamics, tempo variations, and subtle timing shifts. Other attempts, such as ASAP~\cite{foscarin2020asap} with human-assisted beat correction, and advancements in automatic beat tracking, such as Beat This!~\cite{foscarin2024beatthisaccuratebeat}, aim to bring annotation to live-recorded datasets. However, accuracy limitations still present challenges.

\subsubsection{Computational Limitations}

Although using MIDI datasets recorded from live performances and music studios allows for the generation of music with rich and expressive details, generating such music over long durations remains a challenge. This is mainly due to the substantial increase in computational resources required for processing long-range contextual dependencies. Almost all current models, e.g., Music Transformer, employ strategies such as chunking and quantization to reduce token sequence length and vocabulary size~\cite{huang2018musictransformer}. While these approaches help to reduce computational burdens, they also limit the model’s ability to capture ultra-long dependencies and compromise expressive performance details. As stated on the Music Transformer webpage,\footnote{\url{https://magenta.tensorflow.org/music-transformer}} ``Some `failure’ modes include too much repetition, sparse sections, and jarring jumps.” This trade-off prevents the effective generation of long-term coherent symbolic music.

\subsection{Capturing Long-term Coherence}

In the efforts to learn musical structures and generate symbolic music with long-distance coherence, the approaches can generally be categorized into two main types, i.e., those that utilize \textit{explicit} structural features and those that encourage the model to learn \textit{implicit} structural features. Each approach is introduced in the following subsections. 

\subsubsection{Explicit Structural Features}

The explicit use of structural features often relies on handcrafted feature engineering and external analysis tools. A common method in these models is a waterfall-like approach. Typically, the process begins by generating a lead sheet and then using the lead sheet as a condition for the subsequent generation tasks.

MuseMorphose~\cite{wu2022musemorphosefullsongfinegrainedpiano} explicitly controls rhythmic intensity and polyphony density on a bar basis by training on datasets with bar annotations, specifically the LPD-17-cleansed and Pop1K7 datasets. Compose \& Embellish~\cite{wu2023composeembellishwellstructured} leverages third-party tools such as the skyline algorithm, edit similarity, and A* search measures to extract melody and identify structural patterns, enabling the model to produce music with enhanced structural organization. However, it still relies on bar-annotated datasets for training. Rule-Guided Diffusion~\cite{huang2024symbolicmusicgenerationnondifferentiable} uses note density and chord to condition generation, resulting in structured musical segments. Its training does not rely on an annotated dataset but rather uses the performance dataset, Maestro. Still, it only produces short musical pieces instead of full-length segments. Whole-Song Hierarchical Generation~\cite{wang2024wholesonghierarchicalgenerationsymbolic} is capable of generating fully structured, complete pieces of music. It employs a multi-stage approach, using annotations from the POP909 dataset~\cite{wang2020pop909popsongdatasetmusic}, including chord information and separate tracks for melody, bridge, and accompaniment, to produce structured elements such as form, lead sheet, and accompaniment, ultimately generating a complete, full-length piece.

\subsubsection{Implicit Structural Features}

Another important approach involves enabling the model to learn the structural information of music implicitly. The method has the advantage of being more generalizable, as it does not rely on handcrafted feature engineering. However, its downside lies in the increased difficulty for the model to capture complex structural features of music.

MusicVAE~\cite{roberts2019hierarchicallatentvectormodel} uses a bidirectional RNN and a Conductor RNN to generate per-bar latent vectors that are decoded into individual notes, focusing on bar-level structure through training on datasets containing bar annotations, which may not be applicable to freely performed music. Museformer~\cite{yu2022museformertransformerfinecoarsegrained} applies sparse Transformer attention by fully attending to all tokens in selected bars and the summarized vectors of all preceding bars, allowing it to capture long-range context with limited computational resources. However, it still leverages the Lakh MIDI dataset~\cite{raffel2016learning}’s bar annotations, which cannot be used with unannotated performance datasets.

\subsection{Unconditional Generation for Full-Length Music}

In the field of symbolic music generation, most studies adopt one or both of two primary paradigms: conditional and unconditional generation. For example, Music Transformer~\cite{huang2018musictransformer} employs both approaches. In a narrow sense, conditional generation refers to using a short segment of an existing music sequence as a prompt, which serves as the context for generating new tokens. The model then generates new tokens autoregressively based on this prompt, either by appending them to form a longer sequence or by extracting the generated continuation as the final output.

However, in such cases, the result is either not fully generated by the model if the prompt is included in the output, or, if only the continuation beyond the prompt is used, the generated segment lacks a clear and intentional beginning. This limitation becomes particularly problematic in full-length music generation, where human-composed works typically exhibit deliberate harmonic and melodic planning from a well-defined starting point. As highlighted in the work by Dai et al.~\cite{dai2024interconnections}, the section structure is not merely an arbitrary construction, but reflects significant interactions with harmony, melody, rhythm, and predictivity. Consequently, unconditional generation, in which the model produces the entire sequence from scratch without any initial prompt, is essential for achieving structurally complete music with a clear beginning and end.

It is worth noting that broader definitions of conditional generation may also include non-prompt-based conditioning or other control signals. However, such cases fall outside the scope of this study, which specifically aims to investigate the model’s ability to generate complete musical works under unconditional settings.

\subsection{SOTA Solutions with Long-Term Dependency on Performance Datasets}

The Perceiver AR model~\cite{hawthorne2022generalpurposelongcontextautoregressivemodeling} from DeepMind has been a significant source of inspiration. It combines cross-attention and self-attention mechanisms, enabling the model to attend to sequences with up to tens of thousands of tokens. Like Perceiver~\cite{jaegle2021perceivergeneralperceptioniterative} and Perceiver IO~\cite{jaegle2022perceiveriogeneralarchitecture}, Perceiver AR~\cite{hawthorne2022generalpurposelongcontextautoregressivemodeling} uses a shorter query in its cross-attention mechanism to attend to much longer sequences, thereby minimizing computational costs. As noted in the paper, this approach allows the model to attend to up to 32,768 tokens when trained on the Maestro dataset~\cite{hawthorne2019enablingfactorizedpianomusic}, offering a significantly longer context compared to models like Transformer-XL~\cite{dai2019transformerxlattentivelanguagemodels}. This ability to efficiently process long-range contextual data is crucial for learning the structure of entire musical pieces.

When using a context length of $32{,}768$, the standard Transformer incurs the following computational and memory costs for self-attention: FLOPs is $\mathcal{O}(L^2 \cdot d) = 32{,}768^2 \times 1024 \approx 1.10 \times 10^{12}$, and memory usage is $\mathcal{O}(L^2) = 32{,}768^2 \approx 1.07 \times 10^9$ attention scores.

In contrast, Perceiver AR applies cross-attention from a much shorter latent sequence, significantly reducing the cost of subsequent self-attention. For example, if the latent length is $N = 1{,}024$, the self-attention within the latent space requires FLOPs of $\mathcal{O}(N^2 \cdot d) = 1{,}024^2 \times 1024 \approx 1.07 \times 10^9$, and memory usage of $\mathcal{O}(N^2) = 1{,}024^2 \approx 1.05 \times 10^6$ attention scores.

This example shows that Perceiver AR can substantially reduce both computational cost and memory usage. Compared to the standard Transformer, the reduction is approximately 1,024 times in computation and 1,024 times in memory consumption.

However, Perceiver AR leverages the last $N$ tokens as the Query with a limited causal mask, and training with teacher-forcing on long sequences led to lower quality generation. Furthermore, we observed that relying solely on long, especially ultra-long, context resulted in repetitive segments in the latter part of the generated content.

Despite significant progress, existing research has yet to fully address the challenge of generating music with both long-term dependencies and expressive performance details, as most approaches either rely heavily on manual processes or face performance limitations due to restricted context window lengths. 

Among the state-of-the-art models discussed above, Perceiver AR stands out as the one that employs the longest token context length in its experiments. It is also trained directly on performance data and does not require annotations such as beat, bar, or phrase. These characteristics make it uniquely well-suited as a baseline for our goal of learning to generate full-length music with structural completeness and expressive nuance from raw performance data.

\section{Preliminaries} \label{sec:preliminaries}

In this section, we introduce the fundamental concepts and key challenges necessary to understand our proposed model. We review the operational mechanism of cross-attention in Perceiver AR, the role of its causal mask, and the key considerations when using ultra-long sequences as context for token generation.

\subsection{Input Sequence pre-processing}

Let the complete sequence be \( \mathbf{X} = \{x_1, x_2, \ldots, x_l\} \), where \( l \) is the total length of the entire music sequence, and \( m \) is the maximum input length, representing the longest sequence that the model can attend to in one pass. The query length is denoted by \( n \), which represents the number of tokens the model uses to query the context, and typically, \( n \leq m \).

In a typical Transformer approach, a segment of length \( m \) is extracted from the sequence at random. Specifically, we select a starting index \( s \in [1, l - m + 1] \), and the segment used for training, denoted as \( \hat{\mathbf{X}} \), is defined in \eqref{eq:transformer_slicing}.

\begin{equation}
\label{eq:transformer_slicing}
\hat{\mathbf{X}} = \{x_s, x_{s+1}, \ldots, x_{s+m-1}\}
\end{equation}

This approach produces overlapping fixed-length windows for training, as defined in \eqref{eq:overlapping_windows}:

\begin{equation}
\label{eq:overlapping_windows}
\{x_1, x_2, \ldots, x_m\}, \{x_2, x_3, \ldots, x_{m+1}\}, \ldots
\end{equation}

These sequences will be handled using a causal mask.

\subsubsection{Causal Masking in Typical Transformers}

In a typical transformer with causal masking, the goal is to ensure that when generating token \( i \), the model does not attend to tokens \( j \) where \( j > i \). The causal mask for this is typically represented by \eqref{eq:causal_mask}:

\begin{equation}
\label{eq:causal_mask}
M_{ij} = \begin{cases} 
0 & \text{if } i \geq j \\
-\infty & \text{if } i < j
\end{cases}
\end{equation}

This matrix is added to the attention scores \( QK^T \) (as defined in \eqref{eq:attention}) so that all positions \( j > i \) (i.e., future tokens) are masked out by setting their attention scores to \( -\infty \), ensuring they don't influence the generation of the current token.

For example, consider a case where the query has a length of \( n = 5 \) and the key/value has a length of \( m = 5 \). The expected causal mask (Vanilla Transformer) is as follows in \eqref{eq:vanilla_causal_mask}:

\begin{equation}
\label{eq:vanilla_causal_mask}
M =
\begin{bmatrix}
0 & -\infty & -\infty & -\infty & -\infty \\
0 & 0 & -\infty & -\infty & -\infty \\
0 & 0 & 0 & -\infty & -\infty \\
0 & 0 & 0 & 0 & -\infty \\
0 & 0 & 0 & 0 & 0
\end{bmatrix}, \quad M \in R^{5 \times 5}
\end{equation}
This mask ensures that the third query token only attends to the first three tokens.

\subsubsection{Perceiver AR's Causal Mask Issue}

In Perceiver AR~\cite{hawthorne2022generalpurposelongcontextautoregressivemodeling}, the situation is different because the query token length \( n \) is much smaller than the key and value lengths \( m \). Specifically, the causal mask only works on the final part of the context sequence, equivalent to the length of the query \( n \), but does not mask tokens that occur before that. This results in some tokens before the query length being visible during training, which is not an issue for generation, except that the segment from \( x_1 \) to \( x_{m-n} \) is not properly learned by the model.

Let’s denote the sequence of keys and values as \( K \) and \( V \), respectively, and the query length as \( n \), while the context length (keys/values) is \( m \), where \( m > n \). The attention mask matrix \( M \) in Perceiver AR can be represented as follows:

\[
M_{ij} = \begin{cases} 
0 & \text{if } j \leq m - n + 1 \\
-\infty & \text{if } j - i > m - n +1 \\
0 & \text{if } i > n 
\end{cases}
\]

Below shows an example of the causal mask used in Perceiver AR with \(n = 5\) (query length) and \(m = 10\) (context length) as illustrated in \eqref{eq:perceiver_causal_mask}:

\begin{equation}
\label{eq:perceiver_causal_mask}
M =
\begin{bmatrix}
0 & 0 & 0 & 0 & 0 & 0 & -\infty & -\infty & -\infty & -\infty \\
0 & 0 & 0 & 0 & 0 & 0 & 0 & -\infty & -\infty & -\infty \\
0 & 0 & 0 & 0 & 0 & 0 & 0 & 0 & -\infty & -\infty \\
0 & 0 & 0 & 0 & 0 & 0 & 0 & 0 & 0 & -\infty \\
0 & 0 & 0 & 0 & 0 & 0 & 0 & 0 & 0 & 0
\end{bmatrix},
\end{equation}
\begin{equation*}
M \in R^{5 \times 10}
\end{equation*}

Thus, the model can ``peek" at tokens before the query length because they are not fully masked, allowing it to attend to tokens before the intended context during training. This partial masking leads to a mismatch between training and auto-regressive generation, degrading the quality of the generated music.

Note that this issue primarily affects unconditional generation, where the model generates music without any specific prompt or primer, making it more sensitive to inconsistencies between training and generation contexts.

In conditional generation, where the model starts with an initial prompt or primer sequence, using a primer with a length close to the unmasked portion during training can help preserve generation quality and prevent degradation.

\subsubsection{Calculation for Attention with Mask}

The attention mechanism with a causal mask is computed as follows~\cite{vaswani2023attentionneed}:

\begin{equation}
\text{Attention}(Q, K, V) = \text{softmax}\left( \frac{Q K^T + M}{\sqrt{d_k}} \right) V
\label{eq:attention}
\end{equation}

\subsection{Ultra-Long Context in auto-regressive Generation}

Similar to the phenomenon of repeated segments often observed in natural language generation as described in \cite{holtzman2020curiouscaseneuraltext}, relying only on the ultra-long context in auto-regressive models can lead to generated sequences containing excessively long repetitive short segments, especially as the sequence length increases. Given that the probabilities of generating tokens $x_t$ and $x_{t-k}$ (where $k$ is a small integer) are, respectively, as shown in \eqref{eq:prob_xt} and \eqref{eq:prob_xtk}:

\begin{equation}
\label{eq:prob_xt}
p(x_t | x_1, x_2, \dots, x_{t-1})
\end{equation}
\begin{equation}
\label{eq:prob_xtk}
p(x_{t-k} | x_1, x_2, \dots, x_{t-k-1})
\end{equation}

As \( t \) increases, the contexts \( [x_1, x_2, \dots, x_{t-k-1}] \) and \( [x_1, x_2, \dots, x_{t-1}] \) become nearly identical due to the ultra-long context. Consequently, the conditional distributions \( p(x_t | x_1, x_2, \dots, x_{t-1}) \) and \( p(x_{t-k} | x_1, x_2, \dots, x_{t-k-1}) \) are almost indistinguishable, resulting in a KL divergence close to zero:

\[
D_{\text{KL}}(p(x_t | x_1, x_2, \dots, x_{t-1}) || p(x_{t-k} | x_1, x_2, \dots, x_{t-k-1})) \approx 0
\]

This similarity in conditional distributions means that the model is likely to generate similar tokens in nearby steps, as the probability distributions governing \( x_t \) and \( x_{t-k} \) become nearly identical. Thus, the probability of \( x_t = x_{t-k} \) increases, leading to repetitive short segments.

When such repetitive tokens are generated across multiple time steps, the sequence exhibits high token autocorrelation. Mathematically, the token autocorrelation at lag \( k \in \{1, 2, \dots, T-1\} \) for the sequence \( \textbf{X} = \{x_1, x_2, \dots, x_T\} \) can be expressed as~\cite{lee2023mathematical}:

\begin{equation}
\label{eq:token_autocorrelation}
\rho_k(\textbf{X}) = \frac{\sum_{t=k+1}^T (x_t - \bar{x})(x_{t-k} - \bar{x})}{\sum_{t=1}^T (x_t - \bar{x})^2}
\end{equation}
where \( \bar{x} \) represents the mean of the sequence \( \textbf{X} \). When values at nearby steps exhibit high similarity, as suggested by nearly identical conditional distributions, the term \( (x_t - \bar{x})(x_{t-k} - \bar{x}) \) in \eqref{eq:token_autocorrelation} becomes large, leading to high token autocorrelation at lag \( k \).

During generation, identical values are not produced at nearby steps to avoid training penalties. In fact, a similar context of neighboring tokens causes the generation process to produce identical or similar tokens at nearby steps, leading to a higher probability of generating repetitive short segments as the sequence grows longer.

In contrast, a short-range context attention mechanism that focuses only on the most recent \( L \) tokens (e.g., \( L = 1024 \)) changes the conditional probability distribution from:

\[
p(x_t | x_1, x_2, \dots, x_{t-1})
\quad \text{to} \quad
p(x_t | x_{t-L}, x_{t-L+1}, \dots, x_{t-1})
\]

This shorter context reduces the overlap between the context of \( x_t \) and that of \( x_{t-k} \). As a result, the KL divergence between their conditional distributions increases:

\[
D_{\text{KL}}(p(x_t | x_{t-L}, \dots, x_{t-1}) \parallel p(x_{t-k} | x_{t-k-L}, \dots, x_{t-k-1})) > 0
\]

This increase in divergence reduces the likelihood that \( x_t \approx x_{t-k} \), thereby lowering the token autocorrelation \( \rho_k(\textbf{X}) \). Specifically, the term \( (x_t - \bar{x})(x_{t-k} - \bar{x}) \) in Equation~\eqref{eq:token_autocorrelation} becomes smaller on average due to decreased similarity between neighboring token predictions. Therefore, the introduction of short-range attention suppresses repetitive short segments and improves local diversity in the generated music.

\section{Proposed Approach}

The proposed model, Perceiver\textit{S}, is a dual approach of Effective \textit{S}egmentation and Multi-\textit{S}cale attention to address the limitations in symbolic music generation. Building on the strengths of Perceiver AR~\cite{hawthorne2022generalpurposelongcontextautoregressivemodeling} and introducing mechanisms to handle both short and ultra-long dependencies, Perceiver\textit{S} (\textit{S}egmentation and \textit{S}cale) achieves greater coherence and expressiveness in generated music.

Since Perceiver AR provides the possibility of accessing extremely long contexts, we attempt to use Perceiver AR as a baseline model to learn entire musical pieces and evaluate the quality of its generation. Our goal for symbolic music generation is to achieve long-term coherence while maintaining diversity within shorter segments. Furthermore, we expect the model to learn patterns similar to human composition, with repetition and development. The detailed steps of our approach and improvements are elaborated in the following subsections. 

\subsection{Improving the Model to Effectively Learn Ultra-Long Sequences}

This section outlines a data pre-processing strategy designed to enhance token generation quality in auto-regressive models. As discussed in the previous section, Perceiver AR’s causal mask has limitations in its coverage of the entire input sequence, making it necessary to implement pre-processing adjustments before feeding data into the model.

We set the maximum context length to \textit{32,768 tokens}. Based on the Perceiver AR's original implementation\footnote{\url{https://github.com/google-research/perceiver-ar}}, we randomly cropped the dataset. In this approach, a random starting point is selected between \textit{0} and \textit{(the sequence length - the maximum input length + 1)}, from which a segment of length equal to the \textit{maximum input length} is taken.

\begin{figure}[ht!]
\centering
\includegraphics[width=0.5\textwidth]{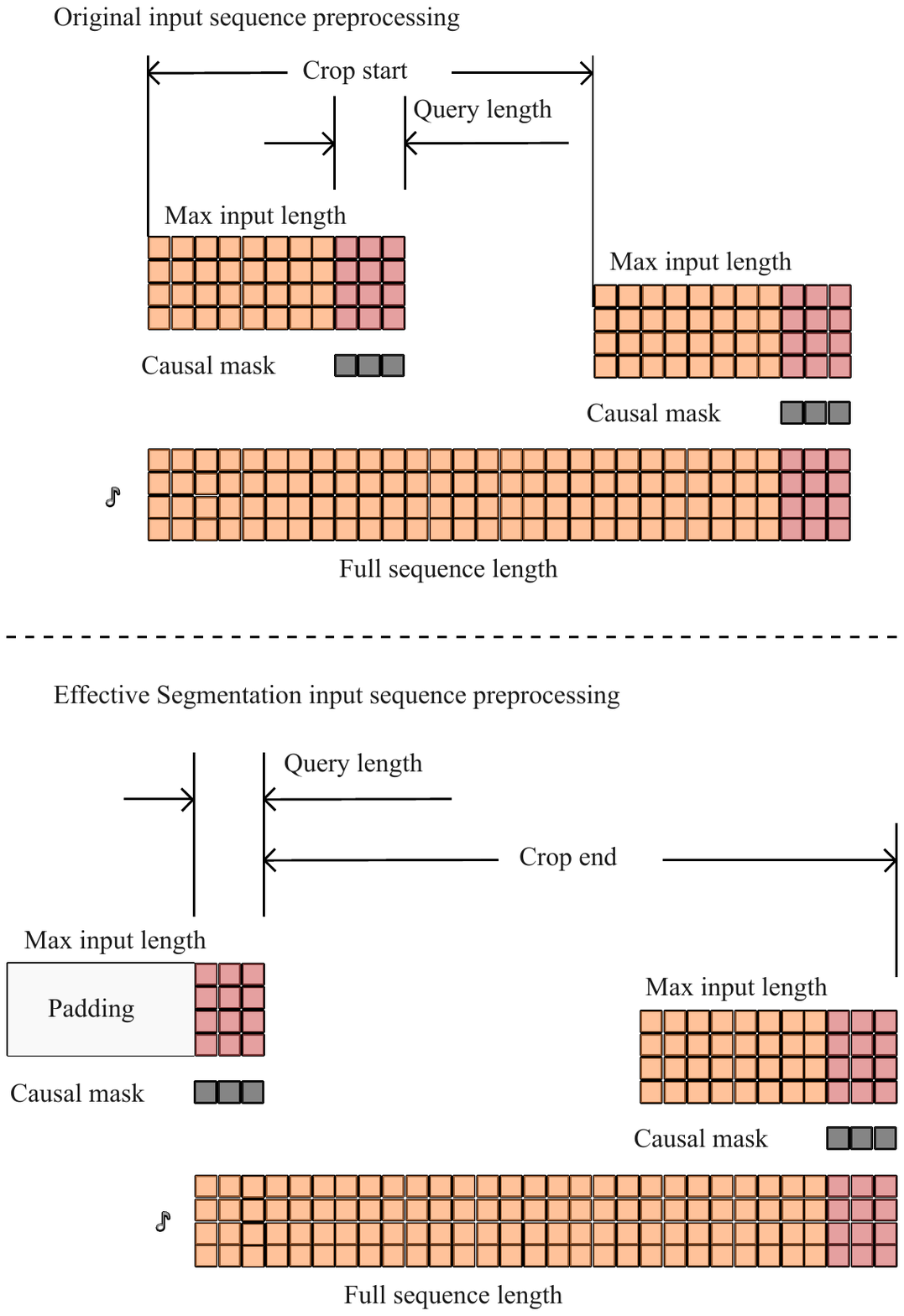}
\caption{Comparison of Data pre-processing Methods and Causal Masks Between the Baseline and Perceiver\textit{S}}
\label{fig:causal_mask}
\end{figure}

The upper part of Figure~\ref{fig:causal_mask} demonstrates the original input sequence pre-processing of the baseline model. Specifically, the baseline model segments the input sequence using the maximum input length as the window size, leaving the beginning of the sequence uncovered by the causal mask. This results in the initial tokens not being predicted or supervised during teacher forcing, preventing the model from learning to generate accurate tokens at the beginning of the sequence due to their exclusion by the causal mask.

In contrast, we propose an alternative method that does not rely on cropping the dataset based on the longest input sequence. Instead, we randomly select an endpoint for cropping between \textit{(the query length + 1)} and \textit{(the sequence length + 1)}. A segment of up to the \textit{maximum input length} is then taken, leading up to this endpoint (or shorter for tokens near the beginning). Padding is applied as in the baseline model.

The lower part of Figure~\ref{fig:causal_mask} illustrates the effective segmentation input sequence pre-processing of the proposed Perceiver\textit{S}. It begins learning token generation from the initial part of the sequence with effective causal mask coverage. This approach ensures that tokens from 1 to  $(m-n)$ in the sequence are effectively covered by the causal mask, enabling the model to learn token generation at the beginning of the sequence more effectively.

The rationale for this improvement is that Perceiver AR~\cite{hawthorne2022generalpurposelongcontextautoregressivemodeling}’s causal mask operates in a \textit{final block} mechanism, where it provides context for the length of the input tokens but only partially masks within the query token length. Training with traditional segmentation causes a mismatch between the training phase, where teacher forcing is used, and the auto-regressive generation process. This approach allows the model to “peek” at unintended tokens during training, which degrades generation quality during inference when such tokens are unavailable for reference.

The experimental results (in the \textbf{Experiments and Results} subsection) will show that this improvement in data preprocessing significantly impacts performance.

\subsection{Further Improving the Model for Generating Music with Both Coherence and Diversity}

After applying ultra-long contexts in the auto-regressive generation, we aim to combine the strengths of both consistency and diversity, allowing the proposed Perceiver\textit{S} to generate music without a tendency toward repeated segments caused by attending solely to long-distance contexts. 

The \textit{Multi-\textit{S}cale} cross-attention mechanism adopted in the proposed Perceiver\textit{S}, while somewhat similar to the concept of Museformer~\cite{yu2022museformertransformerfinecoarsegrained}, is fundamentally different. It introduces multiple layers of attention with varying scales of attending length, designed to balance focus on both long and short contexts, thereby enhancing diversity and reducing repetitive tendencies.

\begin{figure}[ht!]
    \centering
    \includegraphics[width=0.5\textwidth]{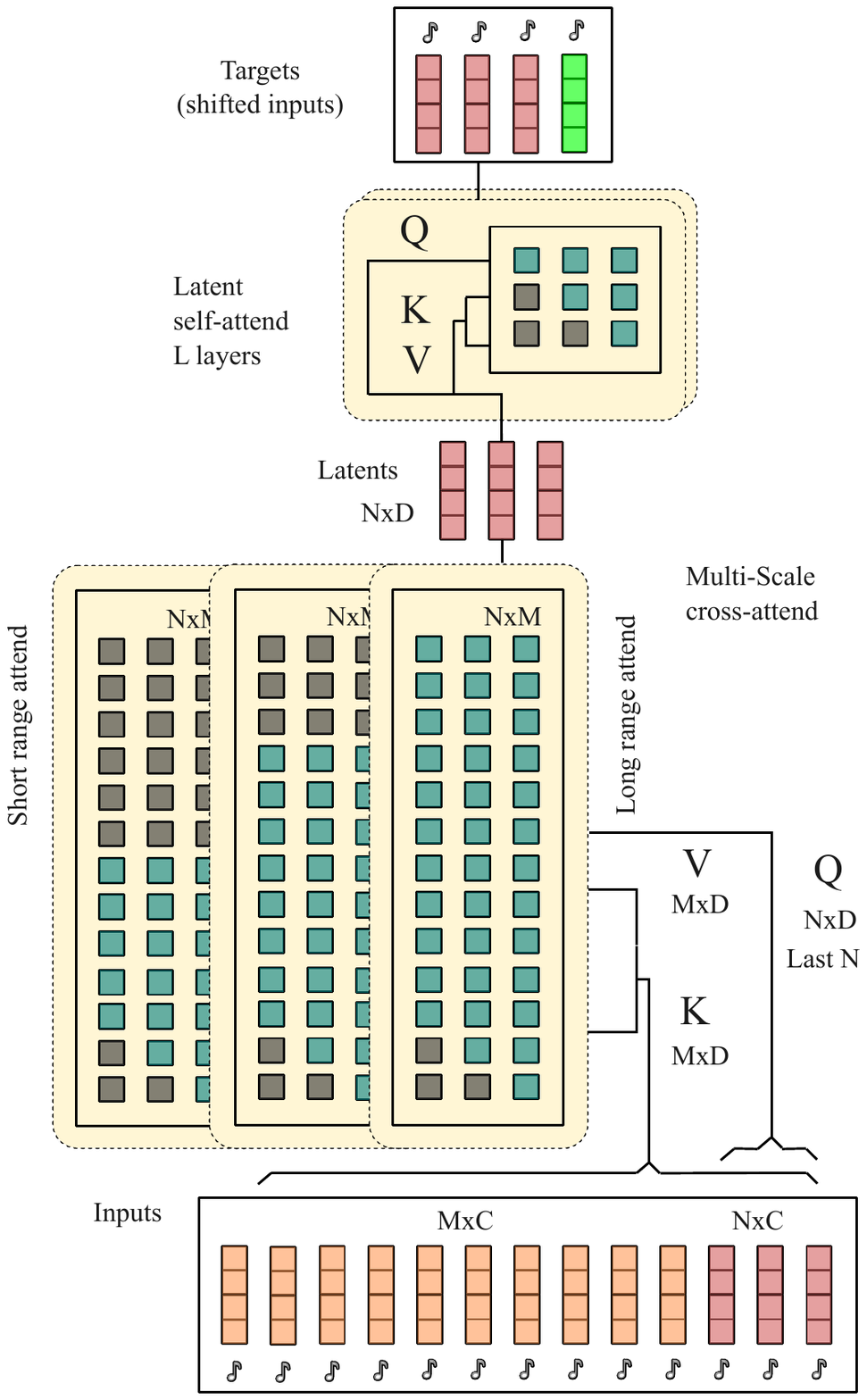}
    \caption{Multi-Scale Causal Cross-Attention Mechanism of Perceiver\textit{S}}
    \label{fig:multi_scale_attention}
\end{figure}

Figure~\ref{fig:multi_scale_attention} illustrates the Multi-Scale Cross-Attention Mechanism of Perceiver\textit{S}. Specifically, within the multi-layer cross-attention, tokens from earlier in the sequence are masked by different scales, and the resulting outputs are combined before being fed into the self-attention layer. This approach enables the model to incorporate cross-attention at multiple scales simultaneously.

Two layers of cross-attention are implemented. One layer operates without a scale mask, while the second layer applies a scale mask that masks out all tokens before the last \textit{1,024 tokens}.

The output of cross-attention layers is combined using the cascade approach. The cascade method feeds the output from the first layer directly as input to the second layer, allowing each layer to refine and build upon the preceding layer’s output.

\section{Technical Details}

In this section, we present the technical details, including the mathematical formulation and technical explanation of our proposed approach, Perceiver\textit{S}, designed for symbolic music generation. Based on the improvements to Perceiver AR, our model introduces two key innovations, i.e., \textit{Effective \textit{S}egmentation} for input sequences and \textit{Multi-\textit{S}cale} cross-attention mechanism. The details are presented in the following sections. 

\subsection{Effective \textit{S}egmentation}

To address the mismatch between training and auto-regressive generation, we propose a novel segmenting method aligned with the causal mask mechanism, whether using random or sequential sampling, that emphasizes shorter context sequences, gradually building up to the maximum input length.

Let the complete sequence be \( \mathbf{X} = \{x_1, x_2, \ldots, x_l\} \), where \( l \) denotes the total length of the entire sequence, \( m \) is the maximum input length the model can attend to in one pass, and \( n \) refers to the query length within that input. The input sequences used for training, denoted as \( \hat{\mathbf{X}} \), are generated as follows in Equation~\eqref{eq:input_sequence_generation}:

\begin{equation}
\label{eq:input_sequence_generation}
\hat{\mathbf{X}}_{i:j} = \{x_i, x_{i+1}, \ldots, x_j\}, \quad \text{where} \ i \leq j \leq i + m - 1
\end{equation}

Here, \( i \) is a starting position chosen within the sequence, and \( j \) is the end position such that the length \( j - i + 1 \) does not exceed the maximum input length \( m \).

When accessing data from the dataset, instead of using fixed-length sequences with the maximum input length, we progressively extract sequences by increasing the context length from \( n \) up to \( m \). This way, the training set includes progressively larger prefixes of the sequence, as shown in Equation~\eqref{eq:progressive_prefixes}:

\begin{equation}
\label{eq:progressive_prefixes}
\begin{aligned}
\{x_1, x_2, \ldots, x_n\}, \{x_1, x_2, \ldots, x_{n+1}\}, \\
\ldots, \{x_1, x_2, \ldots, x_m\}
\end{aligned}
\end{equation}

After reaching the full context length \( m \), we continue segmenting from various starting positions while preserving the maximum input length for each sequence, yielding segments such as in Equation~\eqref{eq:segment_shifting}:

\begin{equation}
\label{eq:segment_shifting}
\begin{aligned}
\hat{\mathbf{X}}_{2:m+1} &= \{x_2, x_3, \ldots, x_{m+1}\}, \\
\hat{\mathbf{X}}_{3:m+2} &= \{x_3, x_4, \ldots, x_{m+2}\}, \\
&\ \vdots
\end{aligned}
\end{equation}

In practice, dataset segments are \textbf{randomly} selected from these defined segments. For \textbf{sequential} segment retrieval, the method begins with progressively longer prefixes, as shown in Equation~\eqref{eq:sequential_prefixes}:

\begin{equation}
\label{eq:sequential_prefixes}
\begin{aligned}
\{x_1, x_2, \ldots, x_{1 \cdot n}\}, & \ \{x_1, x_2, \ldots, x_{2 \cdot n}\}, \\
& \ \ldots, \{x_1, x_2, \ldots, x_{k \cdot n}\} \quad \text{where } k \cdot n \leq m
\end{aligned}
\end{equation}

Here, \( k \) is an integer ranging from 1 to \( K \), where \( K \) represents the maximum number of segments possible within the full sequence length \( l \). When \( kn > m \), subsequent segments shift forward along the sequence in fixed-length windows of size \( m \), as shown in Equation~\eqref{eq:fixed_length_window}:

\begin{equation}
\label{eq:fixed_length_window}
\{x_{k \cdot n - m}, x_{k \cdot n - m + 1}, \ldots, x_{k \cdot n}\}
\end{equation}

This approach ensures that the model learns the sequence structure progressively, closely resembling the auto-regressive generation process, where tokens are predicted one at a time. As a result, the model is better aligned with the auto-regressive nature of music generation, reducing the peeking problem and improving generation quality. Thus, the model can smoothly learn the generation of sequences \( \{x_1, x_2, \ldots, x_{m-n}\} \) with an effective causal mask. The key advantages are summarized as follows.  

\begin{itemize}
    \item \textbf{Progressive learning}: The model learns long-range dependencies progressively by starting with shorter sequences and gradually increasing the context length.
    \item \textbf{Consistency}: This method ensures the training process mimics auto-regressive generation, reducing quality degradation during generation.
    \item \textbf{Causal alignment}: This approach aligns with the causal mask, preventing the model from ``peeking" at future tokens during training.
\end{itemize}

\subsection{Multi-Scale Cross-Attention Mechanism}

The proposed Multi-\textit{S}cale cross-attention mechanism enables Perceiver\textit{S} to handle different levels of context simultaneously. This mechanism employs two cross-attention layers on top of the causal mask: one without any scale mask, allowing all tokens to remain unmasked, and the other layer that masks tokens from the 1st to the \( m-n \)th token. The details with formulas for these two settings are elaborated as follows. 

First, we introduce the \textit{Attention mask without scale mask}. In this case, all tokens are visible to the model, as defined in Equation~\eqref{eq:no_scale_mask}:

\begin{equation}
\label{eq:no_scale_mask}
\hat{M}_{ij} = 0, \quad \forall i, j
\end{equation}

Second, in the case of \textit{Attention mask with scale mask (masking the 1st to the \( m-n \)th tokens)}, tokens from the 1st to the \( m-n \)th positions are masked by the scale mask, preventing the model from attending to these tokens. The scale mask matrix \( \hat{M} \) is defined in Equation~\eqref{eq:scale_mask}:

\begin{equation}
\label{eq:scale_mask}
\hat{M}_{ij} = \begin{cases} 
0 & \text{if } j > m - n \\
-\infty & \text{if } j \leq m - n
\end{cases}
\end{equation}

This ensures that only tokens starting from the \( m-n+1 \)th position are visible, while the model cannot attend to tokens before this position. The combined causal mask \( M \) and scale mask \( \hat{M} \) are added directly to the attention score calculation, modifying the softmax as follows (Equation~\eqref{eq:attention_with_masks}):

\begin{equation}
\label{eq:attention_with_masks}
\text{Attention}(Q, K, V) = \text{softmax}\left( \frac{Q K^T + M + \hat{M}}{\sqrt{d_k}} \right) V,
\end{equation}

\noindent where \( M \) refers to the causal mask that ensures the model does not attend to future tokens, and \( \hat{M} \) denotes the scale mask applied to limit attention to certain parts of the sequence. Both masks work together to control which tokens the model can attend to during training.

After calculating the attention of a given layer in the manner described above, the output from this attention layer is fed as input to the next layer, enabling each layer to refine and build upon the previous layer’s output. We refer to this method of combining attention layers as the \textit{Cascade} mode.

\section{Experiments and Results}

\subsection{Experimental Setup}

In this subsection, we introduce the experimental setup, including dataset selection, MIDI Pre-processing, hyper-parameter selection, and evaluation metrics. 

\subsubsection{Dataset Selection}

In our experiments, we used the Maestro dataset~\cite{hawthorne2019enablingfactorizedpianomusic} as the primary dataset. It contains 1,251 sequences, with a validation set of 240 sequences. All models were trained and evaluated solely on this dataset to ensure consistency in assessing performance.

\subsubsection{MIDI pre-processing}

For MIDI pre-processing, we set the \texttt{Note On} and \texttt{Note Off} events within the range of 0 to 127. \texttt{Time Shift} events were discretized into 100 time steps per second, where each step represents 10 milliseconds. Volume events were quantized into 32 bins, and pedal events were mapped to the duration of relevant notes, discarding the pedal events afterwards. Each song ends with a \texttt{token\_end} marker. No data augmentation, such as key or tempo changes, was applied, though this is planned for future experiments.

\subsubsection{Hyperparameter Selection}

The hidden dimension was set to 1,024 with 24 self-attention layers. Each attention layer had 16 heads, and the head dimension was 64. Adam optimization was used, with an initial learning rate set to 0.03125. Each generated sequence length was set to 8,192 tokens, resulting in approximately 2-10 minutes of music. The music generation in this setup was \textbf{unconditional}, meaning that no external conditions or prompts were used to guide the generation process.

In conditional generation, if the prompt length equals or exceeds the model’s maximum context window, the quality degradation typically caused by the limitations of the causal mask disappears. However, our goal is to generate full-length music with a natural beginning as part of a complete structure through unconditional generation. Therefore, using prompt-based settings that circumvent the limitations of causal masking does not align with the objectives of this study.

It is worth noting that the original Perceiver AR paper \cite{hawthorne2022generalpurposelongcontextautoregressivemodeling} briefly mentions the configuration used for long-context symbolic music generation as “1024 latents and 24 self-attention layers on input contexts of 32,768 tokens.” Beyond this, no further implementation details are disclosed. To ensure a fair comparison in this study, we adhere strictly to these disclosed parameters across both the baseline and improved models. Specifically, all models use a hidden dimension of 1024, 24 layers of latent self-attention, and 2 layers of cross-attention. The total number of parameters for the baseline and both improved models is matched at 361,621,520.

All models were trained on the Maestro dataset using a batch size of 1, a maximum sequence length of 32,768 tokens, a query length of 1,024, and the Adam optimizer with a learning rate following the Noam scheduler with 4,000 warmup steps. Each model was trained for 101 epochs, using cross-entropy loss for both training and validation. The training process exhibited clear validation loss stabilization and overall loss convergence before completion.

For sampling, we use the same decoding parameters for all models: top-p set to 0.75 and temperature set to 1.3. While reducing temperature can indeed help suppress erratic token generation and improve the musicality of the output, particularly in the baseline model, it comes at a cost. Lower temperature reduces diversity, leading to more repetitive and uniform outputs, a common issue in many current state-of-the-art music generation models. For this reason, we deliberately refrain from relying on artificially low temperatures to compensate for the baseline’s deficiencies. 

Finally, 28 MIDI files generated by each model were rendered into audio using the \textit{Vintage Piano} instrument from Logic Pro for subsequent evaluation.

\subsubsection{Evaluation Metrics}

Inspired by the previous research work \cite{yang2020evaluation}, for evaluation, we constructed a reference dataset by separating a set of pieces from the training dataset (Maestro~\cite{hawthorne2019enablingfactorizedpianomusic}) before model training. Then, we generated an equal number of files using the model to form the generated dataset. We calculated the distances within each dataset and between the generated and reference datasets for the following metrics:

\begin{itemize}
    \item \textbf{Total Used Pitch (PC):} Measures the overall pitch diversity by counting distinct pitch classes used throughout the entire piece.
    \item \textbf{Total Used Note (NC):} Counts the distinct notes (pitch and octave combinations) used across the entire piece, indicating the variety in pitch and register.
    \item \textbf{Total Pitch Class Histogram (PCH):} A histogram representing the frequency distribution of pitch classes across the entire piece, offering insights into pitch class preference.
    \item \textbf{Pitch Class Transition Matrix (PCTM):} A matrix representing the probabilities of transitioning from one pitch class to another. This metric captures melodic and harmonic movement patterns.
    \item \textbf{Pitch Range (PR):} Measures the difference between the highest and lowest pitches used in the piece, indicating the overall range of pitches.
    \item \textbf{Average Pitch Interval (PI):} The average interval between consecutive pitches, which reflects the tendency towards stepwise or leapwise motion in melodies.
    \item \textbf{Average Inter-Onset Interval (IOI):} Measures the average time interval between note onsets, capturing the rhythmic density across the entire piece.
    \item \textbf{Note Length Histogram (NLH):} A histogram representing the distribution of note lengths, giving insights into note duration diversity.
    \item \textbf{Note Length Transition Matrix (NLTM):} A matrix representing the transition probabilities between different note lengths, capturing rhythmic patterns and variations in note duration.
\end{itemize}

We introduced four metrics based on time segments, as bar annotations are unavailable in performance datasets, making time-based segmentation essential for evaluating long-term coherence and local diversity.

\begin{itemize}
    \item \textbf{Segment Used Pitch (PC/seg):} Similar to Total Used Pitch, but calculated within segments. This metric helps capture pitch diversity across shorter sections.
    \item \textbf{Segment Used Note (NC/seg):} Counts distinct notes within fixed-length segments, allowing us to analyze the variety of notes used across different parts of the piece.
    \item \textbf{Segment Pitch Class Histogram (PCH/seg):} Represents the pitch class distribution within each segment, providing insights into the consistency of pitch class usage across different sections.
    \item \textbf{Segment Average Inter-Onset Interval (IOI/seg):} Measures the average inter-onset interval within each segment, helping to analyze rhythmic density and tempo consistency within shorter sections.
\end{itemize}

To calculate these additional segment-based metrics, each evaluation piece was divided into 64 segments of equal duration. All of these metrics are calculated based on the \textbf{inter-set} distribution similarity between the generated dataset and the ground truth dataset, including two values: the KL Divergence (KLD) and Overlap Area (OA). OA refers to the overlap between two probability density functions (PDFs), defined as the integral of the pointwise minimum of the two distributions. This value ranges from 0 (no overlap) to 1 (complete overlap), and it quantifies the similarity between two distributions. When the OA value is larger and the KLD value is smaller, it indicates that the above metric distributions of the generated dataset and the ground truth dataset are closely aligned, suggesting that the model has generated music more closely resembling human-composed and performed pieces.

\subsection{Experiments and Results}

Two experiments have been conducted to evaluate the models' performance in various scenarios.

\subsubsection{\textbf{Experiment 1: Input Sequence Segmentation}}
The first experiment aims to demonstrate that Effective \textit{S}egmentation is essential for the model to fully leverage ultra-long-distance context.

We compared two types of input sequence segmentation, with a maximum sequence length set to 32,786 tokens:

\begin{itemize}
    \item Baseline Model: A random starting position is selected within the range [0, sequence length - max input length + 1], and then a segment of max input length tokens is taken from this start.
    \item Improved Model: A random end position is selected within the range [query length + 1, sequence length + 1], and then a segment of tokens, up to the max input length, is taken backwards from this end.
    Segments shorter than the max input length are padded at the beginning. 
\end{itemize}

The segmentation methods produced very different results in auto-regressive training and generation, with the improved model showing much better quality compared with the baseline, as shown in the data within the red dashed box in Table~\ref{tab:result_comparison}. The music generated using the improved input sequence processing approach is closer to the ground truth in terms of lower KLD metrics, including PC, PC/seg, NC/seg, PCH/seg, PCTM, PR, PI, IOI, and IOI/seg, as well as higher OA metrics, including PC, PC/seg, NC, NC/seg, PCTM, PR, PI, IOI, IOI/seg, and NLTM. This result indicates that the improvement effectively enables the model to utilize ultra-long-distance context for music generation.

\begin{table*}[ht!]
\centering
\caption{Evaluation Metrics for Baseline, Effective Segmentation, and Multi-Scale Models}
\label{tab:result_comparison}
\begin{tabular}{ld{3}d{3}d{0}d{3}d{3}d{0}d{3}d{3}d{0}}
\toprule
& \multicolumn{3}{c}{Baseline} & \multicolumn{3}{c}{Effective Segmentation} & \multicolumn{3}{c}{Multi-Scale} \\
\cmidrule(lr){2-4} \cmidrule(lr){5-7} \cmidrule(lr){8-10}
& {KLD} & {OA} & {Comparison} & {KLD} & {OA} & {Comparison} & {KLD} & {OA} & {Comparison} \\
\midrule
PC  & \tikz[overlay, remember picture] \node (start5) {}; 0.036 & 0.614 & 0\% & \tikz[overlay, remember picture] \node (start6) {}; 0.007 & 0.827 & \textbf{+35\%} & 0.017 & 0.759 & \textbf{+24\%} \\
PC/seg  & 0.308 & 0.235 & 0\% & 0.055 & 0.369 & \textbf{+57\%} & 0.055 & 0.704 & \textbf{+200\%} \\
NC  & 0.110 & 0.027 & 0\% & 0.209 & 0.040 & \textbf{+49\%} & 0.165 & 0.040 & \textbf{+49\%} \\
NC/seg  & 0.578 & 0.109 & 0\% & 0.519 & 0.169 & \textbf{+54\%} & 0.626 & 0.151 & \textbf{+38\%} \\
PCH  & 0.017 & 0.948 & 0\% & 0.098 & \tikz[overlay, remember picture] \node (start1) {}; 0.779 & -18\% & 0.060 & \tikz[overlay, remember picture] \node (start3) {}; 0.907 & -4\% \\
PCH/seg  & 0.122 & 0.817 & 0\% & 0.091 & 0.699 \tikz[overlay, remember picture] \node (end1) {}; & -14\% & 0.082 & 0.871 \tikz[overlay, remember picture] \node (end3) {}; & \textbf{+7\%} \\
PCTM  & 0.393 & 0.302 & 0\% & 0.253 & 0.655 & \textbf{+117\%} & 0.177 & 0.424 & \textbf{+41\%} \\
PR  & 0.058 & 0.696 & 0\% & 0.004 & 0.888 & \textbf{+28\%} & 0.013 & 0.774 & \textbf{+11\%} \\
PI  & 0.026 & 0.521 & 0\% & 0.018 & 0.883 & \textbf{+70\%} & 0.071 & 0.729 & \textbf{+40\%} \\
IOI  & 0.327 & 0.670 & 0\% & 0.056 & \tikz[overlay, remember picture] \node (start2) {}; 0.877 & \textbf{+31\%} & 0.004 & \tikz[overlay, remember picture] \node (start4) {}; 0.922 & \textbf{+38\%} \\
IOI/seg  & 0.128 & 0.713 & 0\% & 0.064 & 0.722 & \textbf{+1\%} & 0.038 & 0.860 & \textbf{+21\%} \\
NLH  & 0.057 & 0.841 & 0\% & 0.088 & 0.655 & -22\% & 0.092 & 0.748 & -11\% \\
NLTM  & 0.122 & 0.029 \tikz[overlay, remember picture] \node (end5) {}; & 0\% & 0.187 & 0.040 \tikz[overlay, remember picture] \node (end2) {}; \tikz[overlay, remember picture] \node (end6) {}; & \textbf{+41\%} & 0.139 & 0.048 \tikz[overlay, remember picture] \node (end4) {}; & \textbf{+69\%} \\
\bottomrule
\end{tabular}\\
\vspace{0.2cm}
\footnotesize{The Comparison column shows the percentage difference between the current OA and the baseline OA.}
\begin{tikzpicture}[overlay, remember picture]
    \draw[thick, blue] ($(start1.north west)+(0,0.15)$) rectangle ($(end1.south east)+(0,0.03)$);
    \draw[thick, blue] ($(start2.north west)+(0,0.15)$) rectangle ($(end2.south east)+(0,0.03)$);
    \draw[thick, blue] ($(start3.north west)+(0,0.15)$) rectangle ($(end3.south east)+(0,0.03)$);
    \draw[thick, blue] ($(start4.north west)+(0,0.15)$) rectangle ($(end4.south east)+(0,0.03)$);
    \draw[thick, red, dashed] ($(start5.north west)+(-0.06,0.19)$) rectangle ($(end5.south east)+(0.06,-0.03)$);
    \draw[thick, red, dashed] ($(start6.north west)+(-0.06,0.19)$) rectangle ($(end6.south east)+(0.06,-0.03)$);
\end{tikzpicture}
\end{table*}

\paragraph*{\textbf{Further Training of the Baseline Model under Different Hyperparameters}}
In the above experiment, we trained the baseline model using our current configuration (maximum input length of 32,768, query length of 1,024, and the conventional segmentation approach that randomly selects a starting point and crops a segment of max input length from the original sequence). We observed that the baseline model produced musically incoherent and chaotic outputs. To further investigate the behavior of the baseline model, we trained it under various maximum input length settings, specifically using 1,024, 2,048, 4,096, 8,196, and 16,384 tokens, in addition to the original 32,768-token setup. We then compared these six baseline models against our improved model that incorporates the proposed segmentation strategy.

\begin{figure*}[htbp]
\centering
\includegraphics[width=\textwidth]{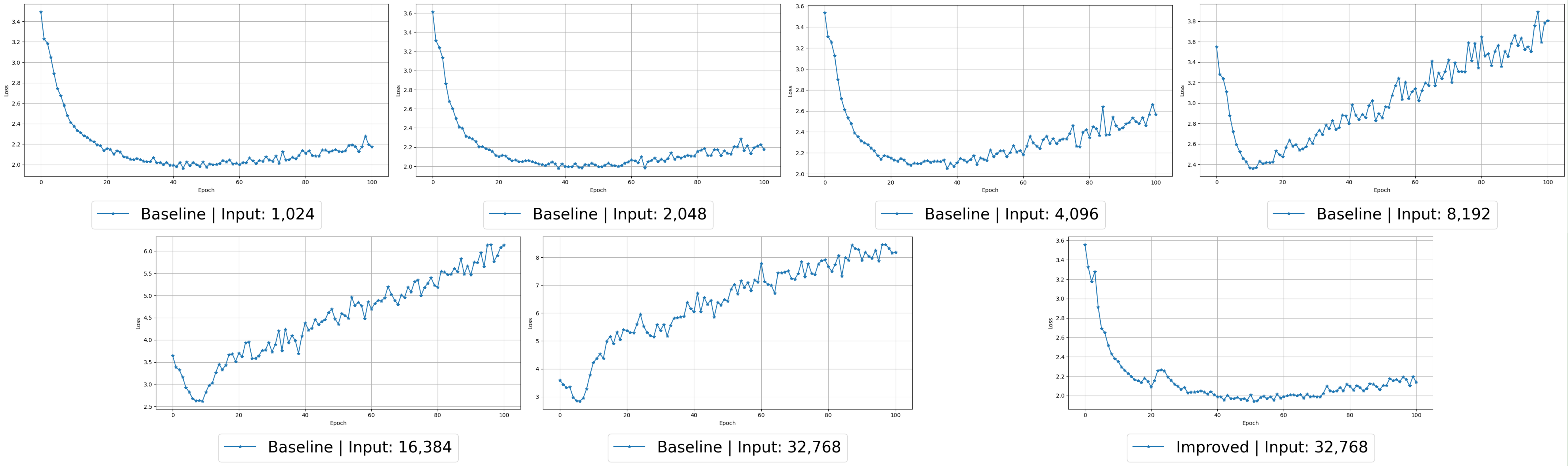}
\caption{Validation losses of baseline models trained with different maximum input lengths. A larger gap between input length and query length leads to slower or failed convergence.}
\label{fig:baseline_val_loss_comparison}
\end{figure*}

By examining the validation loss curves of all models, we observed a consistent trend: the larger the gap between max input length and query length, the more difficult it became for the model’s validation loss to converge (see Figure~\ref{fig:baseline_val_loss_comparison}). This result supports our earlier hypothesis that as the difference between the maximum input length and query length increases, the portion of tokens not covered by the causal mask and therefore not contributing to the loss during training also increases. This results in a higher ratio of untrained tokens, which hampers the model’s ability to generalize. Notably, under the extreme setting of a maximum input length of 1,024 and 2,048, the validation loss exhibited a decreasing trend, demonstrating that our baseline training procedure is itself valid and correctly implemented, and that the model’s degraded performance under other settings is not due to a flawed training methodology but rather to architectural and segmentation mismatches.

To further quantify this effect, we measured the proportion of tokens in the MAESTRO dataset~\cite{hawthorne2019enablingfactorizedpianomusic} that could not be effectively utilized for training under different configurations. The total number of trainable tokens was calculated as the sum of the lengths of all token sequences in the dataset. For each max input length setting, we computed the number of unusable tokens per sequence as \textit{min(max input length, sequence length) - query length}, representing the portion of tokens excluded from contributing to the model’s learning due to the causal attention constraint. The ratio between the excluded tokens and the total tokens was then used to estimate the inefficiency caused by the model design. As shown in Table~\ref{tab:token_efficiency_matrix}, the greater the difference between max input length and query length, the smaller the proportion of tokens that contribute to learning, corroborating our earlier claim that baseline models struggle to learn when the effective training signal is limited. These findings further support our proposed segmentation strategy as a necessary improvement for enabling better model generalization.

\begin{table*}[ht]
\centering
\caption{Token Utilization Efficiency across Query and Max Input Lengths}
\label{tab:token_efficiency_matrix}
\begin{tabular}{rrrrrrr}
\toprule
\multicolumn{7}{c}{\textbf{Cell = Non-Contribution Tokens (Ineffective Token Rate)}} \\
\multicolumn{7}{c}{\textbf{Columns = Max Input Length}} \\
\textbf{Query Length} & \textbf{1,024} & \textbf{2,048} & \textbf{4,096} & \textbf{8,192} & \textbf{16,384} & \textbf{32,768} \\
\midrule
\textbf{1,024}   & 0 (0.00\%) & 1,151,107 (4.45\%) & 3,435,859 (13.28\%) & 7,683,665 (29.71\%) & 13,891,909 (53.71\%) & 20,554,122 (79.47\%) \\
\textbf{2,048}   & 0 (0.00\%) & 0 (0.00\%) & 2,284,752 (8.83\%) & 6,532,558 (25.26\%) & 12,740,802 (49.26\%) & 19,403,015 (75.02\%) \\
\textbf{4,096}   & 0 (0.00\%) & 0 (0.00\%) & 0 (0.00\%) & 4,247,806 (16.42\%) & 10,456,050 (40.43\%) & 17,118,263 (66.19\%) \\
\textbf{8,192}   & 0 (0.00\%) & 0 (0.00\%) & 0 (0.00\%) & 0 (0.00\%) & 6,208,244 (24.00\%) & 12,870,457 (49.76\%) \\
\textbf{16,384}  & 0 (0.00\%) & 0 (0.00\%) & 0 (0.00\%) & 0 (0.00\%) & 0 (0.00\%) & 6,662,213 (25.76\%) \\
\textbf{32,768}  & 0 (0.00\%) & 0 (0.00\%) & 0 (0.00\%) & 0 (0.00\%) & 0 (0.00\%) & 0 (0.00\%) \\
\bottomrule
\end{tabular}\\
\vspace{0.5em}
\footnotesize{\textit{Note:} Each cell shows the number of non-contributing tokens (excluded due to causal mask limits) and the corresponding ineffective token rate in parentheses. The total number of tokens in the Maestro dataset~\cite{hawthorne2019enablingfactorizedpianomusic} is 25,864,128.}
\end{table*}

Given this analysis, the degraded generation quality of the baseline model under the (32,768, 1,024) setting becomes understandable. The original Perceiver AR paper~\cite{hawthorne2022generalpurposelongcontextautoregressivemodeling} also found that increasing the context length on the Wikitext-103 dataset, especially to 4,096 and 8,192, led to degraded perplexity, which is consistent with the results of our experiments. To address this, we propose that segmentation-aware training is especially crucial when working with ultra-long contexts.

\subsubsection{\textbf{Experiment 2: Multi-Scale Attention}}

The aim of the second experiment is to evaluate whether using multi-layer cross-attention with different context scales, especially by simply adding attention outputs for shorter-range dependencies, can improve generation quality.

\begin{figure}[ht!]
\centering
\includegraphics[width=0.45\textwidth]{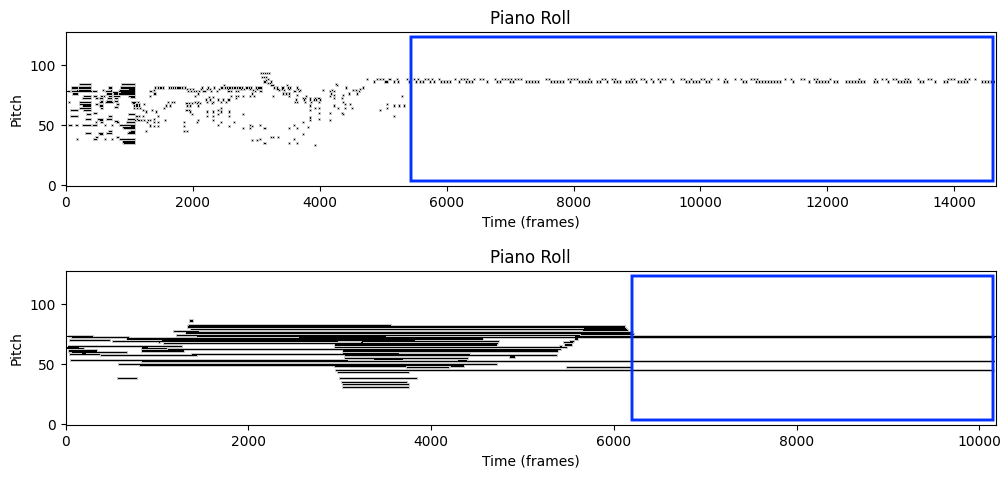}
\caption{Piano roll showing repetitive segments over time in the long context model.}
\label{fig:piano_roll_before_multi_scale}
\end{figure}

While an ultra-long context provides long-term consistency, it tends to generate repetitive segments in the latter part of the sequence (see Figure~\ref{fig:piano_roll_before_multi_scale}). To address this issue, we added different \textbf{scale masks} to the multi-layer cross-attention, allowing different layers to focus on distinct context lengths and merging these results. Here, we define two scale masks for the maximum sequence length of 32,786 tokens. Specifically, in the \textit{No Scale Mask} setting, all tokens remain unmasked. In the \textit{Masked Scale setting}, only the last 1,024 tokens are visible, while all preceding tokens are masked.

The cross-attention layer without a mask is fed into the cross-attention layer with a context mask as its input. We refer to this approach as cascade Multi-\textit{S}cale cross attention.

\begin{figure}[ht!]
\centering
\includegraphics[width=0.45\textwidth]{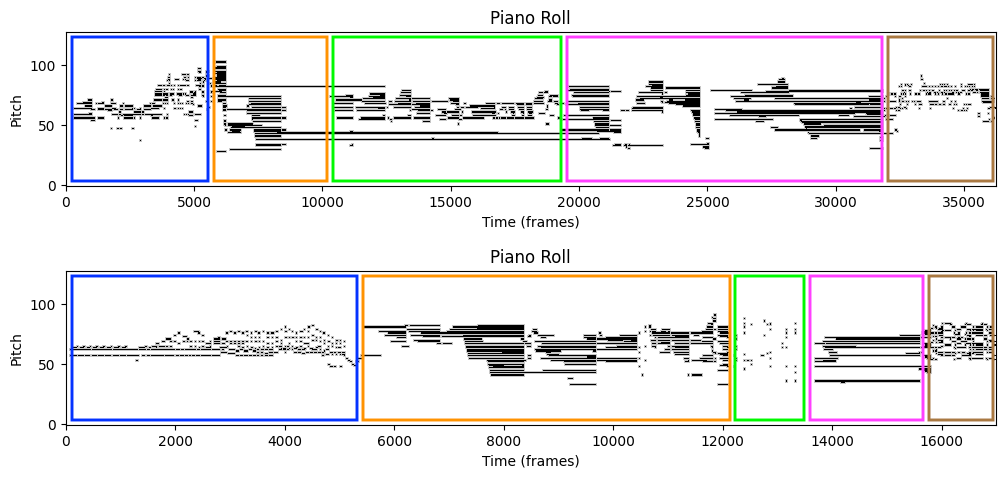}
\caption{Piano roll showing diverse phrases over time in the long context model after applying multi-scale attention.}
\label{fig:piano_roll_after_multi_scale}
\end{figure}

\begin{figure}[htbp]
\centering
\includegraphics[width=0.5\textwidth]{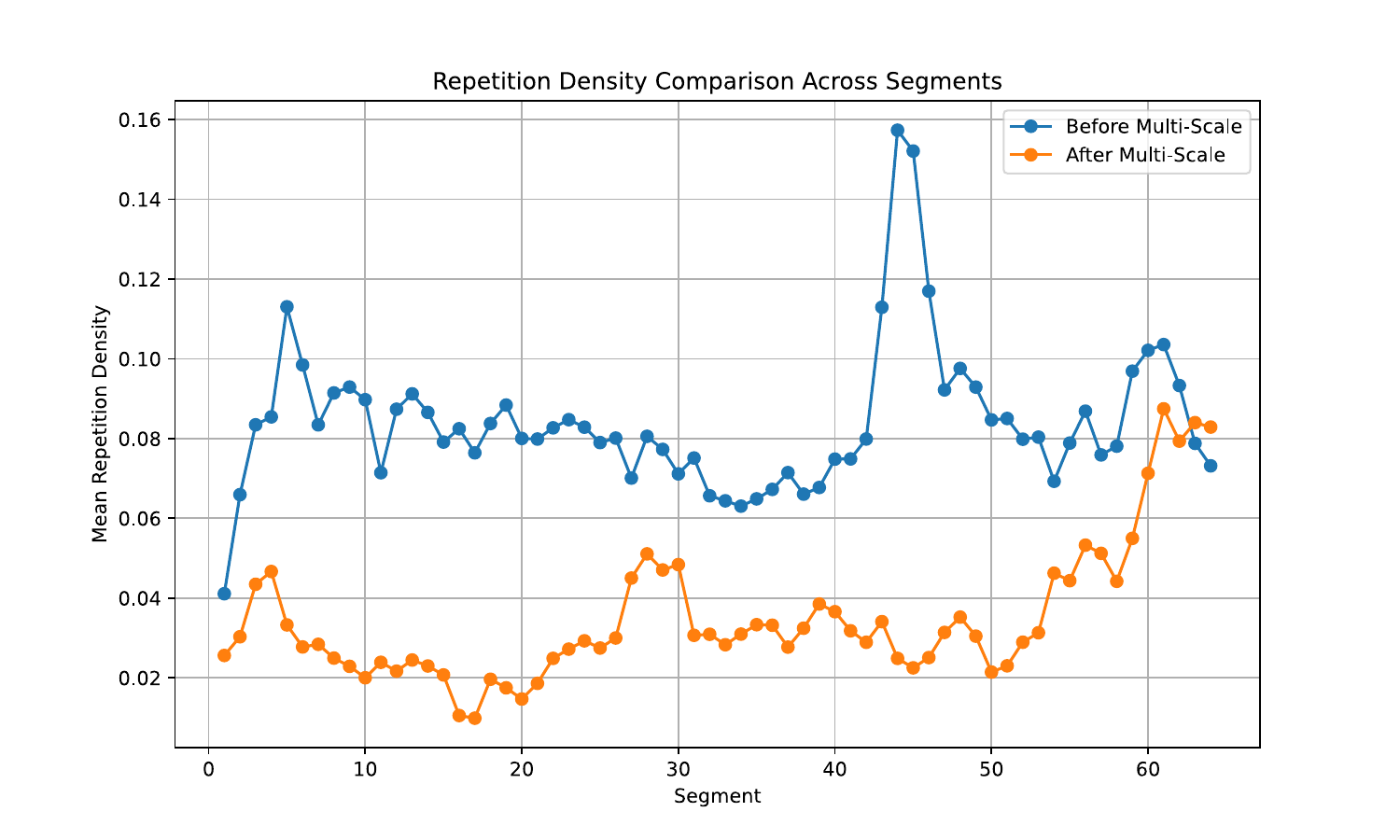}
\caption{Density of repetitive notes detected per unit time in segmented generated music before and after the improvement.}
\label{fig:repetitive_note_density}
\end{figure}

Results, as shown in the data within the solid blue box in \textbf{Table~\ref{tab:result_comparison}}, indicate that this approach significantly improved the model's generation quality. By examining the OA of the four parameters IOI, IOI/seg, NLH, and NLTM—which are related to rhythm characteristics—it is evident that the model's generation aligns more closely with the reference set. Additionally, the OA values for PCH and PCH/seg indicate that the harmonic characteristics also better resemble those of the reference dataset. Furthermore, we analyzed the density of repetitive notes detected within each of 64 equal-length segments (per piece), across all generated samples, both before and after the proposed improvements, as shown in Figure~\ref{fig:repetitive_note_density}, which indicates that the pre-improvement model exhibited a greater tendency to generate repetitive segments. In contrast, unusual repetitive segments in the generated music are now rare after the improvement, as shown in Figure~\ref{fig:piano_roll_after_multi_scale}, with the generated music even exhibiting a rich diversity of phrases.

We also note that the baseline model shows better results in PCH and NLH metrics because most classical music is multi-movement; as a result, multi-movement classical pieces and poorly generated, random-like music, when analyzed across entire pieces, tend to exhibit a wide distribution of pitch classes and note lengths. In contrast, the improved model generates music that maintains relatively more consistent harmony and rhythm throughout an entire piece, resembling a single movement of a classical piece, resulting in a narrower distribution than both ground truth and unstructured music. Therefore, when examining the segmented data for the same metrics on the same set of pieces, such as PCH/seg, the improved model performs better, as the segmented analysis of ground truth music, with most segments taken from the same movement, tends to have a narrower harmonic and rhythmic distribution compared to unstructured music. Although NLH segment data was not calculated, it should follow a similar reasoning.

\begin{table*}[ht]
\centering
\caption{Expert Evaluation Results Based on the Hernández-Oliván et al. Questionnaire}
\label{tab:human_evaluation}
\begin{tabular}{lcccccccc}
\toprule
& \multicolumn{1}{c}{Tempo} & \multicolumn{1}{c}{Motif} & \multicolumn{1}{c}{Rhythm} & \multicolumn{1}{c}{Harmony} & \multicolumn{1}{c}{Style Identified} & \multicolumn{1}{c}{Tonal} & \multicolumn{1}{c}{Human Composed} & \multicolumn{1}{c}{Overall} \\
& Coherence & Presence & Irregularity & Progression & Yes (\%) & Yes (\%) & Yes (\%) & Quality \\
\midrule
Baseline   & 2.86 & 2.14 & 2.71 & 1.64 & 14\% & 4\%  & 18\% & 2.00 \\
PerceiverS & 3.82 & 2.82 & 2.00 & 2.79 & 36\% & 50\% & 29\% & 3.29 \\
\bottomrule
\end{tabular}\\
\vspace{0.5em}
\footnotesize{\textit{Note:} Scores range from 1 (poor) to 5 (excellent). Percentages indicate the proportion of evaluators who selected the affirmative option in each multiple-choice question.}
\end{table*}

Finally, we invited two professional pianists to evaluate the musical outputs generated by the baseline model and our proposed improved model, PerceiverS. Each model produced 14 pieces for evaluation. The evaluation followed the methodology introduced in the questionnaire from Hernández-Oliván et al.’s work on the Subjective Evaluation of Deep Learning Models for Symbolic Music Composition~\cite{hernandezolivan2022subjectiveevaluationdeeplearning}. It was conducted using a blind listening approach and incorporated both quantitative and qualitative assessments. The questionnaire consisted of a 5-point Likert scale evaluating aspects such as tempo coherence, presence of motifs, rhythmic irregularity, harmonic progression, and overall musical quality. In addition, participants were asked to identify the musical style, assess tonality, and determine whether the piece was composed by a human or generated by AI, following a Turing test format.

The evaluation results are summarized in Table~\ref{tab:human_evaluation}, where all criteria confirm through human assessment that PerceiverS significantly outperforms the baseline in generated music quality.

\subsection{Discussion}

Our proposed Effective \textit{S}egmentation technique, which optimizes the input sequence processing for Perceiver AR~\cite{hawthorne2022generalpurposelongcontextautoregressivemodeling}, has shown that, with appropriate data segmentation, Perceiver AR~\cite{hawthorne2022generalpurposelongcontextautoregressivemodeling} can effectively learn long contexts and produce coherent, contextually rich content. The ultra-long context attention indeed provides the model with remarkable benefits in maintaining content consistency over extended generation.

However, a potential drawback of solely relying on long-range dependencies emerged: as sequences get longer, the model increasingly tends to generate repetitive short segments. This occurs because, in the later stages of generation, the long context windows are nearly identical, differing by only a few tokens. Consequently, the probability of generating identical or similar segments increases, further amplified by the high token autocorrelation~\cite{lee2023mathematical} tendency due to the high similarity of context as the sequence lengthens.

To mitigate this issue, we introduce a Multi-\textit{S}cale cross-attention mechanism that combines both long and short context windows. By integrating varying context lengths into multi-layer cross-attention, our approach successfully reduces the tendency for repetitive sequences while maintaining the model’s long-term consistency. This combination strategy enables the generation of high-quality symbolic music over extended sequences.

Our enhanced model, Perceiver\textit{S} (\textit{S}egmentation and \textit{S}cale), leverages Effective \textit{S}egmentation aligned with Perceiver AR~\cite{hawthorne2022generalpurposelongcontextautoregressivemodeling}’s operation mode. By applying Multi-\textit{S}cale masking strategies within multi-layer cross-attention, Perceiver\textit{S} effectively generates cohesive, consistent music over extended temporal contexts, preserving intricate musical patterns and expressive details across long sequences. Additionally, through the use of performance music datasets, Perceiver\textit{S} is capable of producing high-quality symbolic music that captures the nuances of human performance. Importantly, because the model does not rely on annotated datasets, Perceiver\textit{S} can be trained on MIDI data derived from audio using any automatic music transcription (AMT) technique. This capability suggests a future in which symbolic music generation can leverage vast historical recordings, unbounded by the limitations of manually annotated data.

In essence, Perceiver AR~\cite{hawthorne2022generalpurposelongcontextautoregressivemodeling} and, by extension, Perceiver\textit{S}, are general-purpose models adaptable to a wide range of AI tasks. The Effective \textit{S}egmentation and Multi-\textit{S}cale innovations introduced here open up avenues for future applications across domains such as text, image, and video. Future research could thus extend the potential of Perceiver\textit{S}, exploring its capabilities across diverse modalities and expanding its utility within the broader landscape of AI tasks.

\section{Conclusion and Future Work}

In this work, we introduced a novel model, Perceiver\textit{S}, which builds on the Perceiver AR~\cite{hawthorne2022generalpurposelongcontextautoregressivemodeling} architecture by incorporating Effective \textit{S}egmentation and a Multi-\textit{S}cale attention mechanism. The Effective \textit{S}egmentation approach progressively expands the context segment during training, aligning more closely with auto-regressive generation and enabling smooth, coherent generation across ultra-long symbolic music sequences. The Multi-\textit{S}cale attention mechanism further enhances the model's ability to capture both long-term structural dependencies and short-term expressive details.

By addressing limitations in existing models, particularly the issue of causal masking in auto-regressive generation and the high token autocorrelation problem in ultra-long sequences, Perceiver\textit{S} enables the effective handling of ultra-long token sequences without compromising the quality of generated music. Through our proposed Effective \textit{S}egmentation in dataset pre-processing and Multi-\textit{S}cale attention modifications, we demonstrated significant improvements in generating coherent and diverse musical pieces.

Our approach to symbolic music generation provides a new balance between structural coherence and expressive diversity, setting a foundation for future advancements in symbolic music generation models.

\bibliographystyle{unsrt}
\bibliography{ref}

\begin{thebibliography}{10}

\bibitem{liu2023audioldmtexttoaudiogenerationlatent}
Haohe Liu, Zehua Chen, Yi~Yuan, Xinhao Mei, Xubo Liu, Danilo Mandic, Wenwu Wang, and Mark~D. Plumbley.
\newblock Audioldm: Text-to-audio generation with latent diffusion models, 2023.

\bibitem{copet2024simplecontrollablemusicgeneration}
Jade Copet, Felix Kreuk, Itai Gat, Tal Remez, David Kant, Gabriel Synnaeve, Yossi Adi, and Alexandre Défossez.
\newblock Simple and controllable music generation, 2024.

\bibitem{li2023jen1textguideduniversalmusic}
Peike Li, Boyu Chen, Yao Yao, Yikai Wang, Allen Wang, and Alex Wang.
\newblock Jen-1: Text-guided universal music generation with omnidirectional diffusion models, 2023.

\bibitem{hawthorne2022generalpurposelongcontextautoregressivemodeling}
Curtis Hawthorne, Andrew Jaegle, Cătălina Cangea, Sebastian Borgeaud, Charlie Nash, Mateusz Malinowski, Sander Dieleman, Oriol Vinyals, Matthew Botvinick, Ian Simon, Hannah Sheahan, Neil Zeghidour, Jean-Baptiste Alayrac, João Carreira, and Jesse Engel.
\newblock General-purpose, long-context autoregressive modeling with perceiver ar, 2022.

\bibitem{hawthorne2019enablingfactorizedpianomusic}
Curtis Hawthorne, Andriy Stasyuk, Adam Roberts, Ian Simon, Cheng-Zhi~Anna Huang, Sander Dieleman, Erich Elsen, Jesse Engel, and Douglas Eck.
\newblock Enabling factorized piano music modeling and generation with the maestro dataset, 2019.

\bibitem{lee2023mathematical}
Minhyeok Lee.
\newblock A mathematical interpretation of autoregressive generative pre-trained transformer and self-supervised learning.
\newblock {\em Mathematics}, 11(11):2451, 2023.

\bibitem{huang2018musictransformer}
Cheng-Zhi~Anna Huang, Ashish Vaswani, Jakob Uszkoreit, Noam Shazeer, Ian Simon, Curtis Hawthorne, Andrew~M. Dai, Matthew~D. Hoffman, Monica Dinculescu, and Douglas Eck.
\newblock Music transformer, 2018.

\bibitem{kong2022giantmidipianolargescalemididataset}
Qiuqiang Kong, Bochen Li, Jitong Chen, and Yuxuan Wang.
\newblock Giantmidi-piano: A large-scale midi dataset for classical piano music, 2022.

\bibitem{zhang2022atepp}
Huan Zhang, Jingjing Tang, Syed~RM Rafee, Simon Dixon, George Fazekas, and Geraint~A Wiggins.
\newblock Atepp: A dataset of automatically transcribed expressive piano performance.
\newblock In {\em Ismir 2022 Hybrid Conference}, 2022.

\bibitem{edwards2023pijama}
Drew Edwards, Simon Dixon, and Emmanouil Benetos.
\newblock Pijama: Piano jazz with automatic midi annotations.
\newblock {\em Transactions of the International Society for Music Information Retrieval}, 2023.

\bibitem{hawthorne2018onsetsframesdualobjectivepiano}
Curtis Hawthorne, Erich Elsen, Jialin Song, Adam Roberts, Ian Simon, Colin Raffel, Jesse Engel, Sageev Oore, and Douglas Eck.
\newblock Onsets and frames: Dual-objective piano transcription, 2018.

\bibitem{raffel2016learning}
Colin Raffel.
\newblock {\em Learning-based methods for comparing sequences, with applications to audio-to-midi alignment and matching}.
\newblock Columbia University, 2016.

\bibitem{foscarin2020asap}
Francesco Foscarin, Andrew Mcleod, Philippe Rigaux, Florent Jacquemard, and Masahiko Sakai.
\newblock Asap: a dataset of aligned scores and performances for piano transcription.
\newblock In {\em International Society for Music Information Retrieval Conference}, pages 534--541, 2020.

\bibitem{foscarin2024beatthisaccuratebeat}
Francesco Foscarin, Jan Schlüter, and Gerhard Widmer.
\newblock Beat this! accurate beat tracking without dbn postprocessing, 2024.

\bibitem{wu2022musemorphosefullsongfinegrainedpiano}
Shih-Lun Wu and Yi-Hsuan Yang.
\newblock Musemorphose: Full-song and fine-grained piano music style transfer with one transformer vae, 2022.

\bibitem{wu2023composeembellishwellstructured}
Shih-Lun Wu and Yi-Hsuan Yang.
\newblock Compose \& embellish: Well-structured piano performance generation via a two-stage approach, 2023.

\bibitem{huang2024symbolicmusicgenerationnondifferentiable}
Yujia Huang, Adishree Ghatare, Yuanzhe Liu, Ziniu Hu, Qinsheng Zhang, Chandramouli~S Sastry, Siddharth Gururani, Sageev Oore, and Yisong Yue.
\newblock Symbolic music generation with non-differentiable rule guided diffusion, 2024.

\bibitem{wang2024wholesonghierarchicalgenerationsymbolic}
Ziyu Wang, Lejun Min, and Gus Xia.
\newblock Whole-song hierarchical generation of symbolic music using cascaded diffusion models, 2024.

\bibitem{wang2020pop909popsongdatasetmusic}
Ziyu Wang, Ke~Chen, Junyan Jiang, Yiyi Zhang, Maoran Xu, Shuqi Dai, Xianbin Gu, and Gus Xia.
\newblock Pop909: A pop-song dataset for music arrangement generation, 2020.

\bibitem{roberts2019hierarchicallatentvectormodel}
Adam Roberts, Jesse Engel, Colin Raffel, Curtis Hawthorne, and Douglas Eck.
\newblock A hierarchical latent vector model for learning long-term structure in music, 2019.

\bibitem{yu2022museformertransformerfinecoarsegrained}
Botao Yu, Peiling Lu, Rui Wang, Wei Hu, Xu~Tan, Wei Ye, Shikun Zhang, Tao Qin, and Tie-Yan Liu.
\newblock Museformer: Transformer with fine- and coarse-grained attention for music generation, 2022.

\bibitem{dai2024interconnections}
Shuqi Dai, Huan Zhang, and Roger~B Dannenberg.
\newblock The interconnections of music structure, harmony, melody, rhythm, and predictivity.
\newblock {\em Music \& Science}, 7:20592043241234758, 2024.

\bibitem{jaegle2021perceivergeneralperceptioniterative}
Andrew Jaegle, Felix Gimeno, Andrew Brock, Andrew Zisserman, Oriol Vinyals, and Joao Carreira.
\newblock Perceiver: General perception with iterative attention, 2021.

\bibitem{jaegle2022perceiveriogeneralarchitecture}
Andrew Jaegle, Sebastian Borgeaud, Jean-Baptiste Alayrac, Carl Doersch, Catalin Ionescu, David Ding, Skanda Koppula, Daniel Zoran, Andrew Brock, Evan Shelhamer, Olivier Hénaff, Matthew~M. Botvinick, Andrew Zisserman, Oriol Vinyals, and Joāo Carreira.
\newblock Perceiver io: A general architecture for structured inputs \& outputs, 2022.

\bibitem{dai2019transformerxlattentivelanguagemodels}
Zihang Dai, Zhilin Yang, Yiming Yang, Jaime Carbonell, Quoc~V. Le, and Ruslan Salakhutdinov.
\newblock Transformer-xl: Attentive language models beyond a fixed-length context, 2019.

\bibitem{vaswani2023attentionneed}
Ashish Vaswani, Noam Shazeer, Niki Parmar, Jakob Uszkoreit, Llion Jones, Aidan~N. Gomez, Lukasz Kaiser, and Illia Polosukhin.
\newblock Attention is all you need, 2023.

\bibitem{holtzman2020curiouscaseneuraltext}
Ari Holtzman, Jan Buys, Li~Du, Maxwell Forbes, and Yejin Choi.
\newblock The curious case of neural text degeneration, 2020.

\bibitem{yang2020evaluation}
Li-Chia Yang and Alexander Lerch.
\newblock On the evaluation of generative models in music.
\newblock {\em Neural Computing and Applications}, 32(9):4773--4784, 2020.

\bibitem{hernandezolivan2022subjectiveevaluationdeeplearning}
Carlos Hernandez-Olivan, Jorge~Abadias Puyuelo, and Jose~R. Beltran.
\newblock Subjective evaluation of deep learning models for symbolic music composition, 2022.

\end{thebibliography}

\end{document}